\documentclass[journal,twoside,web]{ieeecolor}
\usepackage{tmi}
\usepackage{cite}
\usepackage{amsmath,amssymb,amsfonts}
\usepackage{graphicx}
\usepackage{balance}
\usepackage{textcomp}
\usepackage{booktabs}
\definecolor{proposedrowgray}{gray}{0.91}
\usepackage{multirow}
\usepackage{array}
\usepackage{bm}
\usepackage{url}
\usepackage{algorithm}
\usepackage{algpseudocode}
\usepackage{placeins}
\usepackage{float}
\usepackage{stfloats}
\usepackage[switch]{lineno}
\usepackage{etoolbox}

\pretocmd{\thebibliography}{\normalcolor}{}{}

\algrenewcommand\algorithmicrequire{\textbf{Input:}}
\algrenewcommand\algorithmicensure{\textbf{Output:}}
\algrenewcommand\algorithmicindent{1.15em}
\algrenewcommand\alglinenumber[1]{\scriptsize #1:}
\newcommand{\algdashrule}{%
    \leavevmode\leaders\hbox{\rule{2.2pt}{0.25pt}\hspace{1.8pt}}\hfill\kern0pt%
}

\def\BibTeX{{\rm B\kern-.05em{\sc i\kern-.025em b}\kern-.08em
    T\kern-.1667em\lower.7ex\hbox{E}\kern-.125emX}}
\newcommand{\bvg}{BRIDGE}

\newcommand{\xb}{\mathbf{x}_{b}}
\newcommand{\xg}{\mathbf{x}_{g}}
\newcommand{\zg}{\mathbf{z}_{g}}
\newcommand{\zhatg}{\hat{\mathbf{z}}_{g}}
\newcommand{\zb}{\mathbf{z}_{b}}
\newcommand{\fb}{\mathbf{f}_{b}}
\newcommand{\ellb}{\boldsymbol{\ell}_{b}}
\newcommand{\ellfinal}{\boldsymbol{\ell}_{\mathrm{final}}}
\newcommand{\deltag}{\Delta\boldsymbol{\ell}_{g}}
\newcommand{\pfinal}{\mathbf{p}_{\mathrm{final}}}

\begin{document}

\title{BRIDGE: Learning Brain-Gut Metabolic Evidence for Alzheimer's Disease Diagnosis}

\author{
Pengli Zhu,~\IEEEmembership{Member,~IEEE,}
Ruxi Chang,
Tianliang Yao,
Yu Jiang,
Shengyuan Liu,
Yiyang Lin,
Xiaoyi Duan,
Shuang Wu,
Yixuan~Yuan,~\IEEEmembership{Senior Member,~IEEE,}
and the Alzheimer's Disease Neuroimaging Initiative
\thanks{This work was supported by the National Natural Science Foundation of China (No. 62522124 and 82272051), NSFC/RGC Joint Research Scheme N\_CUHK4126/25 and Innovation and Technology Fund Mainland-Hong Kong Joint Funding Scheme MHP/173/24. (Pengli Zhu and Ruxi Chang contributed equally to this work.) (Corresponding authors: Yixuan Yuan, Shuang Wu, and Xiaoyi Duan.)}
\thanks{\sloppy P. Zhu, T. Yao, Y. Jiang, S. Liu, Y. Lin, and Y. Yuan are with the Department of Electronic Engineering, The Chinese University of Hong Kong, Hong Kong, China (e-mail: penglizhu@\allowbreak cuhk.\allowbreak edu.\allowbreak hk; tlyao@\allowbreak link.cuhk.\allowbreak edu.\allowbreak hk; yujiang@\allowbreak cuhk.\allowbreak edu.\allowbreak hk; liushengyuan@\allowbreak link.cuhk.\allowbreak edu.\allowbreak hk; \mbox{yiyanglin}@\allowbreak link.cuhk.\allowbreak edu.\allowbreak hk; yxyuan@\allowbreak ee.cuhk.\allowbreak edu.\allowbreak hk).}
\thanks{\sloppy R. Chang, S. Wu, and X. Duan are with the PET/CT Center, The First Affiliated Hospital of Xi'an Jiaotong University, No. 277 Yanta West Road, Xi'an 710061, China (e-mail: changruxi@\allowbreak xjtu.\allowbreak edu.\allowbreak cn; wushuang0516@\allowbreak xjtufu.\allowbreak edu.\allowbreak cn; duanxyi@\allowbreak yeah.\allowbreak net).}
}

\maketitle

\begin{abstract}
Alzheimer's disease (AD) diagnosis may benefit from gut metabolic information associated with brain--gut interactions, yet existing models largely analyze cerebral glucose metabolism alone.
Although whole-body fluorodeoxyglucose positron emission tomography (FDG-PET) can provide paired brain--gut observations, routine brain examinations rarely include gut imaging, limiting direct use of this information for diagnosis.
In this paper, we present BRIDGE (Brain and Gut Representation Integration via Distribution-Guided Enhancement), a two-stage framework for transferring brain--gut metabolic knowledge to brain-only diagnosis.
Specifically, BRIDGE integrates a conditional brain-to-gut latent transport module and a disease-evidence-guided cerebral pathway to capture subject-specific cross-organ correspondence and disease-relevant cerebral representations.
For the cerebral pathway, a large language model (LLM) helps organize literature-derived descriptors into human-approved semantic anchors that guide feature learning.
Additionally, we introduce a counterfactual-guided residual fusion mechanism, which uses matched and non-matched virtual-gut evidence to supervise a confidence gate and dynamically regulate the contribution of transported cross-organ information.
At inference, BRIDGE exploits complementary gut-related context using brain PET alone, without measured gut PET.
Extensive experiments across public and private cohorts demonstrate that BRIDGE outperforms brain-only and cross-organ baselines, achieving AUC gains of 6.0\% and 5.2\% over its brain-only counterpart for CN/MCI and CN/AD diagnosis, respectively.
Notably, its ability to transfer complementary brain--gut metabolic context while preserving brain PET as the sole inference input highlights its potential for clinically practical diagnosis.
Our code is available at \url{https://github.com/CUHK-AIM-Group/BRIDGE}.
\end{abstract} 


\section{Introduction}
Alzheimer's disease (AD) is a progressive neurodegenerative disorder associated with cognitive decline and altered cerebral glucose metabolism~\cite{scheltens2021alzheimers}. Within this clinical spectrum, mild cognitive impairment (MCI) represents a clinically heterogeneous stage of cognitive decline and, in some individuals, may precede progression to AD. To characterize these metabolic changes, fluorodeoxyglucose positron emission tomography (FDG-PET) non-invasively assesses cerebral glucose metabolism and is widely used to evaluate AD-related cognitive impairment~\cite{zhang2026whitematter,barnard2025fdgpet}. Recent computational methods have advanced FDG-PET-based diagnosis through convolutional, multiview, and radiomics-guided representation learning~\cite{ding2019deep,pan2021misepynet,jiang2025idlr}. However, these approaches largely analyze cerebral metabolism in isolation, leaving complementary peripheral metabolic information underexplored.

\begin{figure}[!t]
	\centering
	\includegraphics[width=\columnwidth]{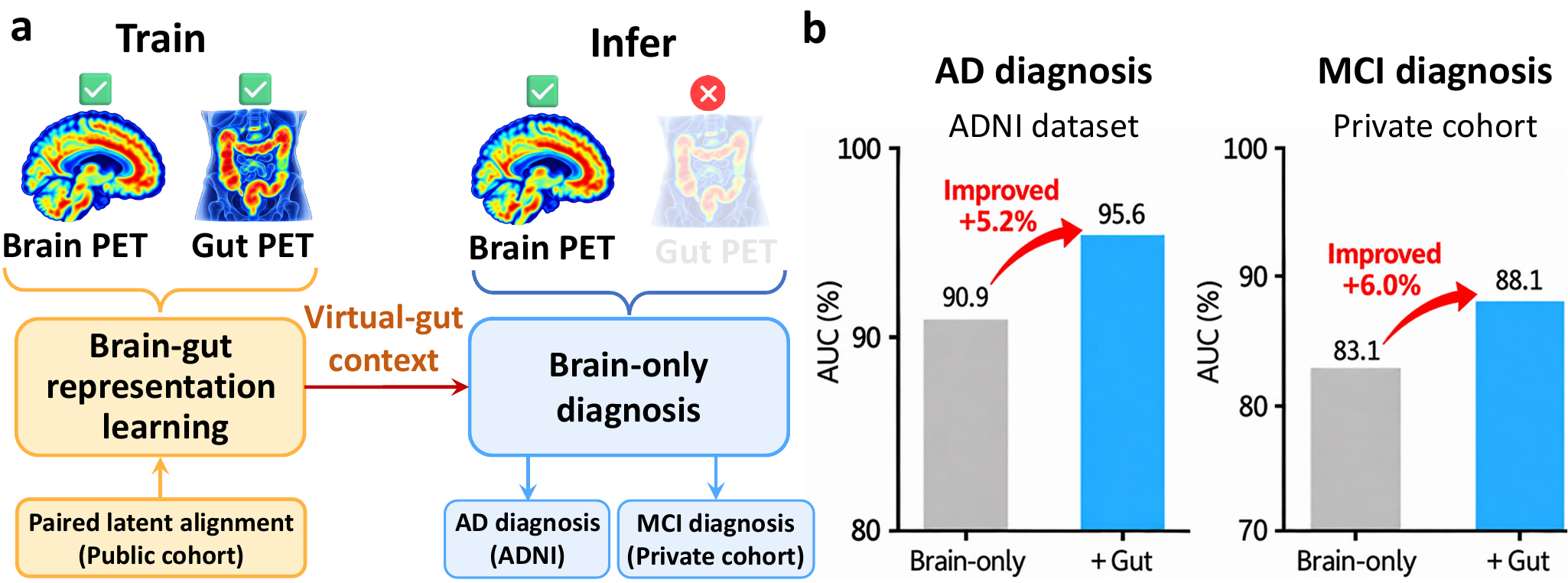}
	\vspace{-17pt}
	\caption{\textbf{Conceptual illustration and performance of \bvg{}.} 
		(a) Brain--gut representation learning and virtual-gut transfer for brain-only inference. 
		(b) AUC comparison between the brain-only and +gut models for AD and MCI diagnosis on the public and private cohorts, respectively.}
	\label{fig:introduction}
	\vspace{-15pt}
\end{figure}

This limitation motivates consideration of systemic metabolic interactions beyond the brain~\cite{arnold2025bioenergetic}. The microbiota--gut--brain axis provides a physiological basis for investigating such associations in AD-related cognitive impairment~\cite{loh2024mgbaxis,ohara2025neuroepithelial,jia2025metagenomic,vemuganti2026imidazole}, while whole-body FDG-PET provides an imaging platform that simultaneously captures cerebral and peripheral glucose metabolism~\cite{geist2025connectome}. Building on this opportunity, recent studies have explored multi-organ representation learning for neurological assessment. BHGCL-guided B-GAT models brain--heart--gut interactions, while multi-constraint knowledge distillation transfers multi-organ information to a deployable brain-only model~\cite{li2026cellreports_bhg,li2026miccai_bhgkd}. MOGAD incorporates inter-organ alignment and hierarchical knowledge transfer, whereas the BHG/Positional framework leverages anatomical prompts and hierarchical multi-organ representations~\cite{zhao2026mogad,li2026tmi_positional}. Collectively, these studies establish the feasibility of using peripheral metabolic information to support neurological diagnosis while retaining brain PET as the sole input at deployment.

Nevertheless, developing a brain-only diagnostic framework that benefits from brain--gut metabolic information at the individual level remains challenging. First, brain and gut PET represent anatomically distinct organs without direct voxel-wise correspondence. Direct brain-to-gut image synthesis is therefore inherently underdetermined, while conventional feature transfer does not explicitly preserve subject-specific correspondence between inferred and measured gut representations. Second, disease-relevant cerebral evidence is not readily available as structured guidance for feature learning. Descriptions of regional uptake patterns and disease stages appear across the literature, while diagnosis labels alone do not explicitly align brain PET features with that evidence. Turning these descriptions into reliable semantic guidance requires literature-grounded organization and human review. Third, transported cross-organ information may not be equally reliable across subjects. Inaccurate inferred representations can introduce misleading evidence and compromise an otherwise reliable cerebral prediction. Therefore, an effective cross-organ framework requires mechanisms to learn subject-specific brain--gut correspondence, organize cerebral features using disease evidence, and assess and regulate the diagnostic contribution of transported information.
 
To address these challenges, we propose \textbf{B}rain and Gut \textbf{R}epresentation \textbf{I}ntegration via \textbf{D}istribution-\textbf{G}uided \textbf{E}nhancement (\bvg{}), a two-stage framework for transferring cross-organ metabolic information to brain-only diagnosis. As illustrated in Fig.~\ref{fig:introduction}, \bvg{} separates brain--gut representation learning from diagnostic learning, allowing paired whole-body PET and labeled brain PET to be utilized across different cohorts. To capture subject-specific cross-organ correspondence without reconstructing gut anatomy, \bvg{} employs conditional brain-to-gut latent transport. For cerebral feature learning, a disease-evidence-guided pathway uses human-approved semantic anchors, organized from the literature with assistance from a large language model (LLM), to structure disease-relevant brain features. Finally, counterfactual-guided residual fusion contrasts matched and non-matched virtual-gut evidence and uses a confidence gate to regulate the contribution of transported information.  

The key contributions are summarized as follows:
\begin{enumerate}
\item We formulate a two-stage cross-organ learning framework that transfers brain--gut metabolic knowledge across disjoint paired whole-body and labeled brain PET cohorts while requiring only brain PET at inference;

\item We introduce conditional latent transport for subject-specific brain--gut correspondence in latent space, together with disease-evidence-guided cerebral learning based on human-approved semantic anchors organized with large language model assistance;

\item We develop a counterfactual-guided residual fusion mechanism that adaptively regulates virtual-gut evidence at inference, and show that the resulting framework improves diagnostic performance over brain-only and cross-organ methods across public and private cohorts.

\end{enumerate}

The paper is organized as follows.
Section~\ref{related} reviews related work on FDG-PET diagnosis, cross-organ learning, and missing-modality transfer. 
Section~\ref{Methodology} presents \bvg{} and its optimization. 
Section~\ref{Study} describes the experimental settings and evaluation results. 
Finally, Section~\ref{CD} concludes this work.

\section{Related Work}\label{related}
\subsection{Brain FDG-PET Diagnosis}
Deep learning has been widely applied to FDG-PET-based diagnosis for characterizing disease-related cerebral metabolic patterns~\cite{sun2025fdgmeta}. Early studies primarily used convolutional networks to extract discriminative PET features~\cite{ding2019deep}, while subsequent work incorporated multiview learning and radiomics-guided representations~\cite{pan2021misepynet,jiang2025idlr}. More recent architectures, including ConvNeXt, Vision Transformers, and Swin Transformers, further improve local and global feature modeling~\cite{woo2023convnextv2,dosovitskiy2021image,liu2021swin}. Semantic guidance has also been explored~\cite{nath2025vilam3,chen2026holodx}, MedCLIP learns from unpaired image--text data, whereas MedKLIP injects entity-level descriptions into local visual representations~\cite{wang2022medclip,wu2023medklip}. Despite these advances, existing approaches predominantly construct diagnostic representations from cerebral imaging alone and do not explicitly incorporate complementary metabolic information from peripheral organs.

\subsection{Cross-Organ Metabolic Learning}
Whole-body PET enables systemic analysis of metabolic relationships across organs and has motivated recent work on multi-organ learning for neurological assessment~\cite{labarthe2026connectomics}. Brain--heart--gut relationships have been modeled through graph-based learning and knowledge distillation~\cite{li2026cellreports_bhg,li2026miccai_bhgkd}. MOGAD introduces inter-organ alignment and hierarchical knowledge transfer~\cite{zhao2026mogad}, while BHG/Positional leverages anatomical prompts for multi-organ learning~\cite{li2026tmi_positional}. These methods demonstrate that peripheral metabolic information can improve neurological assessment while preserving brain-only deployment.

Nevertheless, existing approaches largely rely on joint multi-organ modeling or teacher--student transfer and do not explicitly model subject-specific brain--gut correspondence when paired whole-body PET and labeled brain PET originate from disjoint cohorts. This setting motivates cross-organ knowledge transfer that can learn correspondence independently of downstream diagnostic supervision.

\subsection{Missing-Modality Representation Transfer}
Missing-modality learning exploits complementary information unavailable at inference through shared representations, image synthesis, or knowledge distillation~\cite{han2026imdl,liu2026hilbert,yang2026unimix}. HeMIS and shared-specific modeling accommodate incomplete inputs~\cite{havaei2016hemis,wang2023shaspec}. Image translation methods synthesize missing contrasts from available imaging modalities~\cite{chen2025fdgdiffusion,zhu2026qspace,zhang2025unified}. Knowledge distillation transfers information across modality combinations for reduced-input inference~\cite{kwak2025crossmodal,wang2026e2ad}.

Most existing methods consider modalities or contrasts of the same anatomy, where correspondence is relatively well defined~\cite{zhang2025structure,wang2025unisyn,zhu2025cyclecdm}. Brain and gut PET instead represent anatomically distinct organs without direct voxel-wise correspondence. BRIDGE addresses this setting through representation-level transport that learns subject-specific cross-organ correspondence without anatomical reconstruction and adaptively regulates transported information during diagnosis.

\begin{figure*}[!t]
	\centering
	\includegraphics[width=\textwidth]{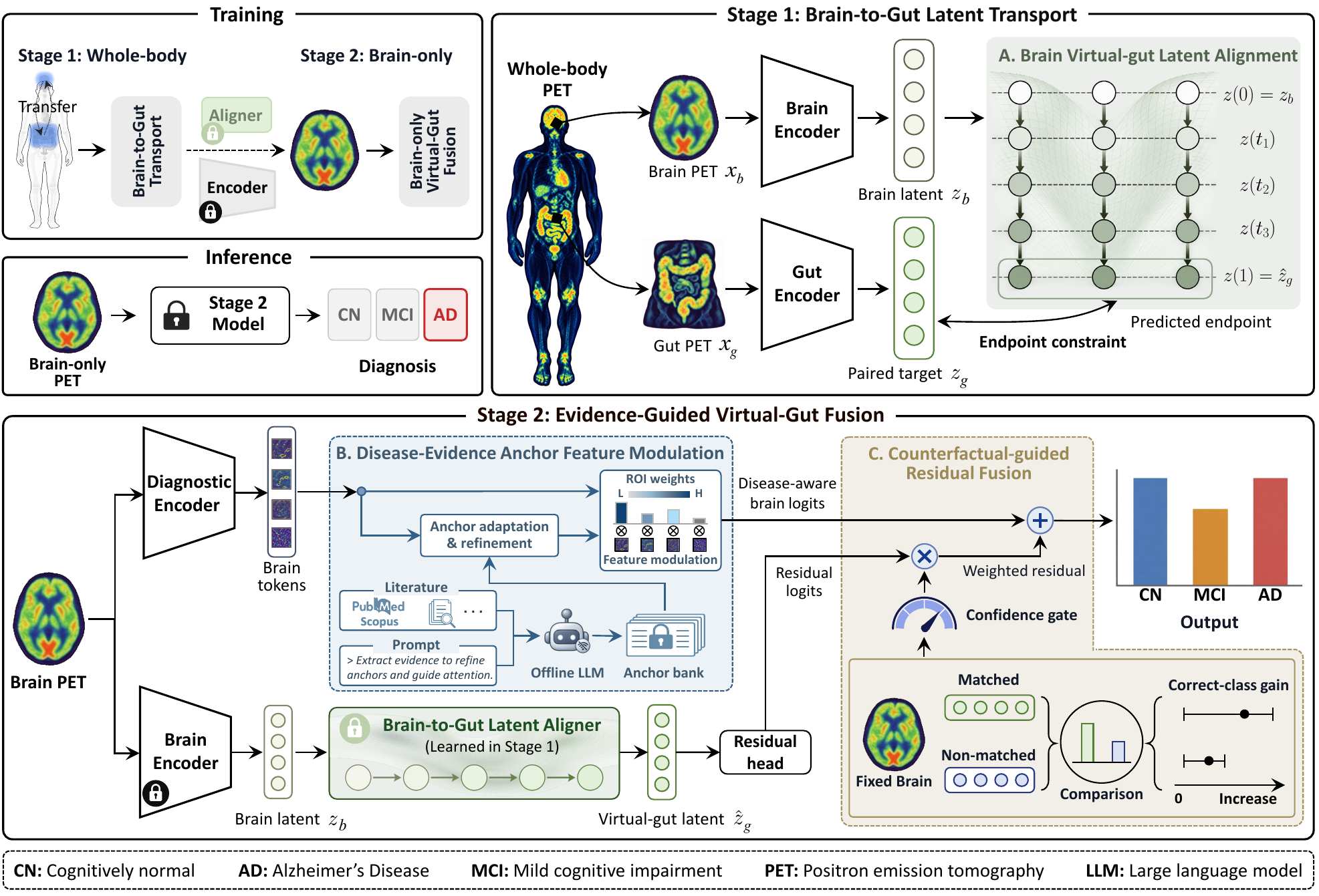}
	\vspace{-15pt}
	\caption{\textbf{Overview of BRIDGE framework.} Stage~1 uses paired whole-body PET to learn subject-specific brain-to-gut latent transport through organ-specific encoders and endpoint constraints. Stage~2 uses labeled brain PET to train a cerebral pathway guided by literature-derived, human-approved semantic anchors and a virtual-gut residual pathway. An offline LLM assists in organizing the anchors, while matched and nonmatched gains supervise the confidence gate. At inference, a brain-only PET drives both pathways, and the gate regulates the virtual-gut contribution.}
	\label{fig:overall}
	\vspace{-10pt}
\end{figure*}

\section{Methodology}\label{Methodology}

\subsection{Overview}
In this paper, we formulate \bvg{} as a two-stage framework for learning subject-specific brain--gut correspondence from paired whole-body PET and transferring it to brain-only diagnosis. The two training cohorts are defined as follows:
\begin{equation}
	\begin{aligned}
		\mathcal{D}_{\mathrm{WB}}
		&=\{(\xb^{i},\xg^{i})\}_{i=1}^{N_{\mathrm{WB}}},\\
		\mathcal{D}_{\mathrm{B}}
		&=\{(\xb^{i},y_i)\}_{i=1}^{N_{\mathrm{B}}},
		\qquad y_i\in\{1,\ldots,C\},
	\end{aligned}
\end{equation}
Here, $\xb^{i}$ and $\xg^{i}$ denote paired brain and gut PET crops, $y_i$ is a diagnostic label among $C$ classes, and $i$ indexes the two cohorts independently. The whole-body cohort $\mathcal{D}_{\mathrm{WB}}$ provides paired observations without requiring diagnostic labels, whereas $\mathcal{D}_{\mathrm{B}}$ provides labeled brain PET without requiring measured gut PET. With these cohorts, our goal is to estimate $p(y\mid\xb)$ by transferring subject-specific brain--gut correspondence learned from $\mathcal{D}_{\mathrm{WB}}$ to brain-only diagnosis on $\mathcal{D}_{\mathrm{B}}$.

As illustrated in Fig.~\ref{fig:overall}, Stage~1 uses paired whole-body PET to learn organ-specific representations and a brain-to-gut transport operator $\mathcal{T}_{\psi}$, with paired endpoint regularization encouraging subject-specific latent correspondence. Stage~2 freezes this operator and combines a cerebral pathway guided by human-approved, LLM-assisted semantic anchors with a virtual-gut residual pathway through a confidence gate supervised during training by the relative diagnostic gains of matched and nonmatched virtual-gut evidence. For a brain scan, the cerebral pathway produces logits $\ellb\in\mathbb{R}^{C}$, while the transported pathway produces residual logits $\deltag\in\mathbb{R}^{C}$. The gate $\alpha\in[0,1]$ controls the contribution of the residual:
\begin{equation}
	\ellfinal=\ellb+\alpha\deltag,
	\quad
	\pfinal=\operatorname{softmax}(\ellfinal).
	\label{eq:final_prediction}
\end{equation}
At inference, both pathways use the same brain scan, and the gate regulates the virtual-gut contribution rather than estimating a calibrated probability of diagnostic correctness.

\subsection{Stage 1: Brain-to-Gut Latent Transport}
To learn subject-specific correspondence between anatomically distinct organs without reconstructing gut anatomy, Stage~1 transports brain representations toward their paired gut representations in a common latent space. To this end, the stage consists of organ-specific encoding, conditional latent transport, and paired endpoint regularization.

\subsubsection{Organ-Specific Latent Encoding}
To avoid imposing a shared anatomical representation on the brain and gut, we use separate three-dimensional encoders $E_b$ and $E_g$ followed by projection heads $P_b$ and $P_g$. They map paired observations into a common $d$-dimensional transport space: $\zb=P_b(E_b(\xb)), \zg=P_g(E_g(\xg)), \zb,\zg\in\mathbb{R}^{d}$. The pairing provides subject identity rather than voxelwise correspondence. The gut encoder and projection define the target representation used to learn transport and are discarded after Stage~1.

\subsubsection{Conditional Cross-Organ Flow}
To model the cross-organ transformation without assuming a direct voxel-level mapping, we learn a continuous latent transport with conditional flow matching~\cite{lipman2023flow,yazdani2025flow,chemseddine2025conditional}. For subject $i$ and $t\sim\mathcal{U}(0,1)$, a straight path connects the paired latents, and the vector field $v_{\psi}$ is trained to recover its target velocity:
\begin{equation}
	\begin{aligned}
		\mathbf{z}_{i,t}
		&=(1-t)\zb^{i}+t\zg^{i},
		\qquad
		\mathbf{u}_{i}=\zg^{i}-\zb^{i},\\
		\mathcal{L}_{\mathrm{FM}}
		&=\mathbb{E}_{i,t}
		\left\|
		v_{\psi}(\mathbf{z}_{i,t},t\mid\zb^{i})-\mathbf{u}_{i}
		\right\|_{2}^{2}.
	\end{aligned}
	\label{eq:flow_matching}
\end{equation}
Conditioning on the source latent $\zb^{i}$ keeps the transported trajectory tied to the input subject. At inference of the transport branch, integrating the learned field from the brain latent yields a deterministic virtual-gut endpoint:
\begin{equation}
	\frac{d\mathbf{z}(t)}{dt}
	=v_{\psi}(\mathbf{z}(t),t\mid\zb),
	\quad
	\mathbf{z}(0)=\zb,
	\quad
	\zhatg=\mathcal{T}_{\psi}(\zb)=\mathbf{z}(1).
	\label{eq:latent_transport}
\end{equation}
This endpoint is a representation-level estimate for downstream diagnosis; it is not intended to reconstruct gut anatomy or model the full conditional distribution of gut states.

\subsubsection{Paired Endpoint Regularization}
To encourage locally learned flow dynamics to reach a subject-consistent gut representation, we explicitly constrain the transported endpoint. Specifically, endpoint fidelity is measured by Euclidean and angular agreement, while a minibatch contrastive term~\cite{oord2018contrastive} favors the paired gut latent over nonmatched subjects:
\begin{equation}
	\begin{aligned}
		\mathcal{L}_{\mathrm{end}}
		&=\mathbb{E}_{i}\!\left[
		\|\zhatg^{i}-\zg^{i}\|_{2}^{2}
		+1-\operatorname{cos}(\zhatg^{i},\zg^{i})\right],\\
		\mathcal{L}_{\mathrm{ctr}}
		&=-\frac{1}{B}\sum_{i=1}^{B}
		\log
		\frac{\exp(\operatorname{sim}(\zhatg^{i},\zg^{i})/\tau)}
		{\sum_{j=1}^{B}\exp(\operatorname{sim}(\zhatg^{i},\zg^{j})/\tau)},
	\end{aligned}
	\label{eq:paired_endpoint}
\end{equation}
where $B$ is the minibatch size, $\operatorname{cos}(\cdot,\cdot)$ and $\operatorname{sim}(\cdot,\cdot)$ denote cosine similarity, and $\tau>0$ is the contrastive temperature. We additionally use the trajectory-energy penalty
$\mathcal{L}_{\mathrm{eng}}
=\mathbb{E}_{i}\int_{0}^{1}
\|v_{\psi}(\mathbf{z}_{i}(t),t\mid\zb^{i})\|_{2}^{2}\,dt$
to discourage unnecessarily large integrated velocities. Therefore, the complete objective for Stage~1 training is formulated as:
\begin{equation}
	\mathcal{L}_{\mathrm{S1}}
	=\mathcal{L}_{\mathrm{FM}}
	+\lambda_{\mathrm{end}}\mathcal{L}_{\mathrm{end}}
	+\lambda_{\mathrm{ctr}}\mathcal{L}_{\mathrm{ctr}}
	+\lambda_{\mathrm{eng}}\mathcal{L}_{\mathrm{eng}}.
	\label{eq:stage1_objective}
\end{equation}
All Stage~1 terms use paired imaging observations only and require no diagnostic labels.

\subsection{Stage 2: Evidence-Guided Virtual-Gut Fusion}
To exploit transported cross-organ context without allowing unreliable virtual-gut evidence to dominate diagnosis, Stage~2 combines an evidence-guided cerebral pathway with a gated virtual-gut residual. The Stage~1 brain encoder, projection, and transport operator remain fixed throughout diagnostic learning.

\subsubsection{Disease-Evidence Anchor Feature Modulation}
To organize cerebral features around disease-relevant metabolic evidence, we use literature-derived descriptors as semantic queries over brain tokens. The diagnostic encoder produces
$\mathbf{T}=E_d(\xb)=[\mathbf{t}_1,\ldots,\mathbf{t}_N]
\in\mathbb{R}^{N\times d}$,
where $N$ is the number of spatial tokens. To provide this guidance, the fixed descriptor bank contains $K=8$ human-approved statements organized offline from literature passages by metabolic pattern, anatomical location, uptake direction, and disease stage. No patient images were provided to the offline LLM. For reproducibility, the archived manifest retains the retrieval date (August 23, 2026), the fixed prompt, four PubMed/PubMed Central queries, seven source records with persistent identifiers and verification status, and the recorded LLM provenance.

Let $\{q_k\}_{k=1}^{K}$ denote these descriptors. Each descriptor is encoded with the BiomedBERT base checkpoint~\cite{gu2022biomedbert} using attention-mask mean pooling, projected from 768 dimensions to $d=256$ by a fixed Gaussian projection with seed 2026, and $\ell_2$ normalized. The complete checkpoint identifier and revision hash are retained in the anchor-embedding artifact. This produces the fixed initialization
$\mathbf{A}^{0}
=[\mathbf{a}^{0}_{1},\ldots,\mathbf{a}^{0}_{K}]
\in\mathbb{R}^{K\times d}$;
a learnable residual gives
$\mathbf{A}=\mathbf{A}^{0}+\Delta\mathbf{A}$.
We refine these anchors with scan-specific brain evidence and use the refined anchors to modulate brain tokens through cross-attention:
\begin{equation}
	\begin{aligned}
		\mathbf{H}
		&=\mathbf{A}+\operatorname{Attn}(\mathbf{A},\mathbf{T},\mathbf{T}),\\
		\widetilde{\mathbf{T}}
		&=\mathbf{T}+\operatorname{Attn}(\mathbf{T},\mathbf{H},\mathbf{H}),\\
		\fb
		&=\Phi_b\!\left(
		[\operatorname{Pool}(\widetilde{\mathbf{T}}),
		\operatorname{Pool}(\mathbf{H})]\right),
		\quad
		\ellb=C_b(\fb).
	\end{aligned}
	\label{eq:anchor_modulation}
\end{equation}
The scan-conditioned anchors $\mathbf{H}$ act as semantic queries rather than anatomical masks, while $\widetilde{\mathbf{T}}$ contains the modulated brain tokens. In Eq.~\eqref{eq:anchor_modulation}, $\operatorname{Pool}$ denotes token pooling, $\Phi_b$ maps the pooled representations to cerebral features, and $C_b$ produces the cerebral logits. To preserve the anchors' literature-derived directions while allowing task adaptation, the cerebral objective augments the cross-entropy classification loss $\mathcal{L}_{\mathrm{CE}}$ with semantic-preservation and diversity regularization:
\begin{equation}
	\begin{aligned}
		\mathcal{L}_{\mathrm{brain}}
		&=\frac{1}{B}\sum_{i=1}^{B}
		\mathcal{L}_{\mathrm{CE}}(\boldsymbol{\ell}_{b,i},y_i)
		+\lambda_{\mathrm{sem}}\mathcal{L}_{\mathrm{sem}}
		+\lambda_{\mathrm{div}}\mathcal{L}_{\mathrm{div}},\\
		\mathcal{L}_{\mathrm{sem}}
		&=\frac{1}{K}\sum_{k=1}^{K}
		\left[
		1-\operatorname{cos}(\mathbf{a}_k,\mathbf{a}^{0}_k)
		\right],\\
		\mathcal{L}_{\mathrm{div}}
		&=\|\bar{\mathbf{A}}\bar{\mathbf{A}}^{\top}
		-\mathbf{I}\|_{F}^{2},
	\end{aligned}
\end{equation}
where each row of $\bar{\mathbf{A}}$ is the $\ell_2$-normalized version of the corresponding row of $\mathbf{A}$, and $\mathbf{I}$ is the $K\times K$ identity matrix. The regularizers encourage semantic preservation and nonredundant anchor directions but do not by themselves guarantee interpretable specialization.

\subsubsection{Fixed Virtual-Gut Evidence Stream}
To expose complementary gut-related context from the same brain scan, the frozen Stage~1 branch maps the input brain PET to a virtual-gut latent, which is converted to a residual correction by a trainable head $C_g$: $\zb=P_b(E_b(\xb)),\zhatg=\mathcal{T}_{\psi}(\zb),
\deltag=C_g(\zhatg)\in\mathbb{R}^{C}$. Accordingly, only $C_g$ is optimized in this stream during Stage~2; the transported representation itself remains fixed. The residual logits are not hard-bounded, and their magnitude is controlled by the fusion objective below.

\subsubsection{Counterfactual-Guided Residual Fusion}\label{sec:confidence_fusion}
To guide residual admission, we compare matched and nonmatched virtual-gut evidence within each diagnostic class and use their diagnostic gains to construct a soft target for the confidence gate. For subject $i$, the cerebral logits are held fixed while its matched virtual-gut latent is replaced by the latent of another record $j(i)\neq i$ from the same diagnostic class:
\begin{equation}
	\begin{aligned}
		\boldsymbol{\ell}^{\mathrm{m}}_{i}
		&=\boldsymbol{\ell}_{b,i}
		+C_g(\hat{\mathbf{z}}_{g,i}),\\
		\boldsymbol{\ell}^{\mathrm{nm}}_{i}
		&=\boldsymbol{\ell}_{b,i}
		+C_g(\hat{\mathbf{z}}_{g,j(i)}).
	\end{aligned}
	\label{eq:counterfactual_logits}
\end{equation}
Within each minibatch, same-class members are randomly permuted and circularly shifted to construct $j(i)$. Samples without a same-class partner are excluded from gate supervision and the matched-gain penalty but remain in the classification and residual-regularization objectives.

Let
$S(\boldsymbol{\ell},y)
=\log[\operatorname{softmax}(\boldsymbol{\ell})]_y$
denote the correct-class log probability. We measure how much the matched and nonmatched residuals improve this evidence relative to the cerebral prediction and convert their difference 
into a soft admission target as follows:
\begin{equation}
	\begin{aligned}
		g_i^{\mathrm{m}}
		&=S(\boldsymbol{\ell}^{\mathrm{m}}_i,y_i)
		-S(\boldsymbol{\ell}_{b,i},y_i),\\
		g_i^{\mathrm{nm}}
		&=S(\boldsymbol{\ell}^{\mathrm{nm}}_i,y_i)
		-S(\boldsymbol{\ell}_{b,i},y_i),\\
		c_i^{*}
		&=\sigma\!\left(
		\frac{
			g_i^{\mathrm{m}}-\max(g_i^{\mathrm{nm}},0)
		}{\tau_g}
		\right).
	\end{aligned}
	\label{eq:confidence_target}
\end{equation}
Thus, the admission target $c_i^{*}$, whose softness is controlled by the positive temperature $\tau_g$, exceeds $0.5$ only when the matched residual provides positive gain and exceeds the corresponding nonmatched gain. This target guides gate learning with diagnostic labels during training, enabling confidence estimation from brain-derived features at inference.

To adaptively determine whether the transported evidence should contribute to
the final prediction, we predict a confidence score from the cerebral feature,
virtual-gut latent, residual logits, and cerebral predictive entropy. Let
$h_i=\mathcal{H}(\operatorname{softmax}(\boldsymbol{\ell}_{b,i}))$ denote the
predictive entropy. The gate and final logits are then computed as:
\begin{equation}
	\begin{gathered}
		\mathbf{r}_i=[\mathbf{f}_{b,i},\hat{\mathbf{z}}_{g,i},\Delta\boldsymbol{\ell}_{g,i},h_i],\\[1pt]
		\alpha_i=\sigma(G_{\omega}(\mathbf{r}_i)),\\[1pt]
		\boldsymbol{\ell}_{\mathrm{final},i}=\boldsymbol{\ell}_{b,i}+\alpha_i\Delta\boldsymbol{\ell}_{g,i}.
	\end{gathered}
	\label{eq:gated_residual}
\end{equation}
where $G_{\omega}$ is a trainable scalar-output gate network and $\sigma$ is the logistic sigmoid. To supervise residual admission, we detach the confidence target as $\bar c_i=\operatorname{stopgrad}(c_i^{*})$ and combine gate supervision with a matched-gain penalty over the valid within-batch replacements $\mathcal{V}$:
\begin{equation}
    \begin{aligned}
    \mathcal{L}_{\mathrm{gate}}
    &=\frac{1}{B}\sum_{i\in\mathcal{V}}
    \operatorname{BCE}(\alpha_i,\bar c_i),\\
    \mathcal{L}_{\mathrm{cf}}
    &=\frac{1}{B}\sum_{i\in\mathcal{V}}[-g_i^{\mathrm{m}}]_{+}.
    \end{aligned}
    \label{eq:gate_supervision}
\end{equation}
where $\operatorname{BCE}$ denotes binary cross-entropy and $[u]_{+}=\max(u,0)$, with both losses set to zero when $\mathcal{V}$ is empty. The matched-gain penalty $\mathcal{L}_{\mathrm{cf}}$ discourages reductions in correct-class evidence relative to the cerebral prediction. Gradients propagate through the predicted gate and matched gain, while the confidence target remains detached. These terms are complemented by final-prediction classification and residual magnitude regularization:
\begin{equation}
	\begin{aligned}
		\mathcal{L}_{\mathrm{cls}}
		&=
		\frac{1}{B}
		\sum_{i=1}^{B}
		\mathcal{L}_{\mathrm{CE}}
		(\boldsymbol{\ell}_{\mathrm{final},i},y_i),\\
		\mathcal{L}_{\mathrm{res}}
		&=
		\frac{1}{B}
		\sum_{i=1}^{B}
		\|\Delta\boldsymbol{\ell}_{g,i}\|_2^2.
	\end{aligned}
	\label{eq:cls_res_loss}
\end{equation}
These terms jointly supervise the fused prediction, guide residual admission, and control residual magnitude through the overall fusion objective:
\begin{equation}
    {\mathcal{L}_{\mathrm{fusion}}
    =\mathcal{L}_{\mathrm{cls}}
    +\lambda_{\mathrm{cf}}\mathcal{L}_{\mathrm{cf}}
    +\lambda_{\mathrm{gate}}\mathcal{L}_{\mathrm{gate}}
    +\lambda_{\mathrm{res}}\mathcal{L}_{\mathrm{res}}.}
	\label{eq:fusion_objective}
\end{equation}
where the weights $\lambda_{\mathrm{cf}}$, $\lambda_{\mathrm{gate}}$, and $\lambda_{\mathrm{res}}$ balance the contributions of the matched-gain penalty, gate-supervision loss, and residual regularization, respectively.

\subsection{Optimization and Brain-Only Inference}
To preserve the separation between cross-organ correspondence learning and disease supervision, we optimize the two stages sequentially. Stage~1 minimizes $\mathcal{L}_{\mathrm{S1}}$ in Eq.~\eqref{eq:stage1_objective} using paired whole-body PET. We then freeze $E_b$, $P_b$, and $\mathcal{T}_{\psi}$ and discard $E_g$ and $P_g$. Before joint Stage~2 training, the cerebral branch is warmed up with $\mathcal{L}_{\mathrm{brain}}$. The diagnostic components are subsequently optimized with
$\mathcal{L}_{\mathrm{S2}}
=\mathcal{L}_{\mathrm{brain}}
+\lambda_{\mathrm{fusion}}\mathcal{L}_{\mathrm{fusion}}$. 

At inference, a single brain PET scan is processed by the diagnostic branch and the frozen transport branch. Their outputs are combined according to Eq.~\eqref{eq:final_prediction}. Accordingly, measured gut PET, diagnostic labels, nonmatched replacement samples, and the offline LLM are not required.

\setlength{\dbltextfloatsep}{10pt plus 2pt minus 2pt}

\begin{table}[H]
\centering
\caption{Cohort composition and experimental roles. BG and B denote paired brain--gut PET and brain PET, respectively. $N$ denotes the number of examinations.}
	\label{tab:data}
	\renewcommand{\arraystretch}{1.3}
\setlength{\tabcolsep}{1pt}
		\begin{tabular*}{\columnwidth}{@{\extracolsep{\fill}}lcccc@{}}
			\toprule
			Cohort & Modality & Labels & $N$ & Role \\
			\midrule
			AutoPET~\cite{gatidis2022autopet} & BG & Unlabeled & 188 & Transport pretrain \\
			PETWB~\cite{xue2026petwbrep} & BG & Unlabeled & 484 & Transport pretrain \\
			QUADRA~\cite{gutschmayer2025quadra} & BG & Unlabeled & 96 & Transport pretrain \\
			ENHANCE~\cite{ferrara2026enhance} & BG & Unlabeled & 12 & Transport pretrain \\
			ADNI~\cite{jack2008alzheimer} & B & CN/MCI/AD & 1,004 & \begin{tabular}[c]{@{}c@{}}Source training\\[-2pt]validation\end{tabular} \\
			Private cohort & BG & CN/MCI & 278 & Primary evaluation \\
			\bottomrule
	\end{tabular*}
	\vspace{-10pt}
\end{table}

\begin{table*}[!h]
\centering
\caption{Private-cohort CN/MCI classification performance (mean$\pm$SD over five folds). B and BG denote brain and brain--gut PET. \\$^*$ indicates significance by two-sided paired $t$-test with Holm correction ($p<0.05$). Best and second-best results are shown in bold and underlined, respectively.}
\label{tab:private_comparison}
\setlength{\tabcolsep}{4.2pt}
\renewcommand{\arraystretch}{1.25}
\small
\begin{tabular*}{\textwidth}{@{\extracolsep{\fill}}lcllllll@{}}
\toprule
\textbf{Method}
& \textbf{{Category}}
& \multicolumn{1}{c}{\textbf{AUC}$\uparrow$}
& \multicolumn{1}{c}{\textbf{ACC}$\uparrow$}
& \multicolumn{1}{c}{\textbf{BACC}$\uparrow$}
& \multicolumn{1}{c}{\textbf{SEN}$\uparrow$}
& \multicolumn{1}{c}{\textbf{SPE}$\uparrow$}
& \multicolumn{1}{c}{\textbf{Macro-F1}$\uparrow$} \\
\midrule

\multicolumn{8}{@{}l}{\textit{Brain-only baselines}} \\
\addlinespace[0.08em]

ConvNeXt~\cite{woo2023convnextv2}
& B/B
& \underline{0.837$\pm$0.050}
& 0.728$\pm$0.040$^*$
& 0.727$\pm$0.039$^*$
& \underline{0.816$\pm$0.149}
& 0.638$\pm$0.101$^*$
& 0.726$\pm$0.041$^*$ \\

3D ViT~\cite{dosovitskiy2021image}
& B/B
& 0.684$\pm$0.070$^*$
& 0.626$\pm$0.068$^*$
& 0.626$\pm$0.068$^*$
& 0.629$\pm$0.090$^*$
& 0.624$\pm$0.091$^*$
& 0.625$\pm$0.069$^*$ \\

Swin-T~\cite{liu2021swin}
& B/B
& \underline{0.837$\pm$0.060}
& 0.756$\pm$0.060
& 0.756$\pm$0.060
& 0.743$\pm$0.092
& 0.769$\pm$0.071
& 0.755$\pm$0.060 \\

MiSePyNet~\cite{pan2021misepynet}
& B/B
& 0.736$\pm$0.055$^*$
& 0.651$\pm$0.075$^*$
& 0.653$\pm$0.075$^*$
& 0.493$\pm$0.203$^*$
& \underline{0.812$\pm$0.132}
& 0.632$\pm$0.095$^*$ \\

\addlinespace[0.16em]
\specialrule{\lightrulewidth}{0pt}{0pt}
\addlinespace[0.16em]
\multicolumn{8}{@{}l}{\textit{Privileged-training comparators}} \\
\addlinespace[0.08em]

MOGAD~\cite{zhao2026mogad}
& BG/B
& 0.792$\pm$0.084$^*$
& 0.724$\pm$0.100$^*$
& 0.724$\pm$0.101$^*$
& 0.721$\pm$0.159
& 0.727$\pm$0.188
& 0.719$\pm$0.104$^*$ \\

BHG/Positional~\cite{li2026tmi_positional}
& BG/B
& 0.814$\pm$0.069$^*$
& 0.723$\pm$0.067$^*$
& 0.722$\pm$0.066$^*$
& 0.786$\pm$0.149
& 0.659$\pm$0.086$^*$
& 0.720$\pm$0.067$^*$ \\

\addlinespace[0.18em]
\specialrule{\lightrulewidth}{0pt}{0pt}
\addlinespace[0.18em]
\multicolumn{8}{@{}l}{\textit{Proposed method}} \\
\addlinespace[0.08em]

\rlap{\textcolor{proposedrowgray}{\rule[-\dimexpr\dp\strutbox+2pt\relax]{\textwidth}{\dimexpr\ht\strutbox+\dp\strutbox+4pt\relax}}}Brain-only
& B/B
& 0.831$\pm$0.030
& \underline{0.766$\pm$0.032}
& \underline{0.767$\pm$0.032}
& 0.764$\pm$0.086$^*$
& 0.769$\pm$0.069
& \underline{0.766$\pm$0.031} \\

\rlap{\textcolor{proposedrowgray}{\rule[-\dimexpr\dp\strutbox+2pt\relax]{\textwidth}{\dimexpr\ht\strutbox+\dp\strutbox+4pt\relax}}}\textbf{\bvg{}}
& BG/B
& \textbf{0.881$\pm$0.050}
& \textbf{0.818$\pm$0.028}
& \textbf{0.819$\pm$0.027}
& \textbf{0.821$\pm$0.067}
& \textbf{0.816$\pm$0.029}
& \textbf{0.815$\pm$0.027} \\
\bottomrule
\end{tabular*}
\end{table*}

\section{Experiments}\label{Study}

\subsection{Setup and Details}\label{sec:experimental_setup}

\subsubsection{Study Cohorts}
Table~\ref{tab:data} summarizes the three complementary data resources used in our experiments. Four public whole-body FDG-PET/CT cohorts, including QUADRA~\cite{gutschmayer2025quadra}, AutoPET~\cite{gatidis2022autopet}, PETWB~\cite{xue2026petwbrep}, and ENHANCE~\cite{ferrara2026enhance}, provide 780 unlabeled paired brain--gut examinations for transport pretraining. ADNI~\cite{jack2008alzheimer} provides 1,004 labeled brain examinations spanning CN, MCI, and AD for source training and evaluation. The private cohort contains 278 paired examinations (138 CN and 140 MCI) for adaptation and held-out CN/MCI evaluation.

Private-cohort PET/CT acquisition followed a 6-h fast and confirmation of blood glucose below 11.1~mmol/L. After intravenous $^{18}$F-FDG administration (3.7~MBq/kg) and 60~min of rest, PET was acquired at eight bed positions (1.5~min each). PET and CT used 5-mm slices and $128\times128$ and $512\times512$ matrices, respectively, with three-iteration time-of-flight PET reconstruction followed by PET/CT fusion. Subsequent cohort selection required cerebral and abdominopelvic coverage, valid PET/CT pairing, and usable CT-derived gut masks, excluding patients with confirmed brain or gastrointestinal tumors. The MCI group additionally required clear signs of age-related cerebral changes. Quality control excluded examinations with inadequate coverage, artifacts, invalid spatial metadata, or failed registration or mask checks. These criteria defined the examination-level cohort used for subsequent model adaptation and held-out CN/MCI evaluation.

\subsubsection{Data Preprocessing}
Brain preprocessing included spatial harmonization, intensity normalization, and quality control. T1-weighted MRI, when available, underwent N4 bias-field correction, affine registration to the MNI template with ANTs, and skull stripping with SynthStrip. Cerebral PET was rigidly registered to the corresponding MRI, transformed to MNI space using the MRI-derived transformation, normalized by pons-referenced uptake, and masked to remove extracranial signal. Registered volumes were visually inspected for anatomical coverage, alignment, and orientation.

Gut preprocessing included CT-based segmentation, PET-space mapping, resampling, and intensity normalization. Gut masks were generated from whole-body CT using TotalSegmentator~\cite{wasserthal2023totalsegmentator} by combining the duodenum, small intestine, colon, sigmoid colon, and rectum, and were mapped to PET space after rigid CT--PET registration. Gut PET was resampled to $1\times1\times1,\mathrm{mm}^3$, while cerebral PET from all cohorts was affinely harmonized to a reference with ANTsPyX and visually checked. Cerebral and gut volumes were resized to $96\times112\times96$ and $96\times64\times128$ voxels, respectively, and normalized using robust scan-wise intensity normalization.

\subsubsection{Implementation Details}
\bvg{} was implemented in Python 3.10 and PyTorch 2.6 on NVIDIA H200 NVL GPUs. Source initialization used 60 epochs, followed by private-cohort fine-tuning for up to 70 epochs with patience 15. The AdamW optimizer used learning rates of $2\times10^{-4}$ and $5\times10^{-5}$ for the two stages, respectively, with weight decay $10^{-4}$ and batch size eight. Mixed precision and gradient clipping at 5.0 were used throughout. The transport solver used eight explicit Euler steps with midpoint vector-field evaluation. For the source cerebral training, $\lambda_{\mathrm{sem}}=0.05$ and $\lambda_{\mathrm{div}}=0.01$, while both weights were zero during private brain-head adaptation. The reported fusion runs used $\lambda_{\mathrm{cf}}=0.2$, $\lambda_{\mathrm{gate}}=0.2$, $\lambda_{\mathrm{res}}=0.005$, and $\tau_g=0.2$ with the counterfactual objective.

\begin{table*}[t]
\centering
\caption{ADNI CN/AD classification performance (mean$\pm$SD over five folds). B and BG denote brain and brain--gut PET. $^*$ indicates significance by two-sided paired $t$-test with Holm correction ($p<0.05$). Best and second-best results are shown in bold and underlined, respectively.}
\label{tab:adni_cnad_folds}
\setlength{\tabcolsep}{4.2pt}
\renewcommand{\arraystretch}{1.25}
\small
\begin{tabular*}{\textwidth}{@{\extracolsep{\fill}}lcllllll@{}}
\toprule
\textbf{Method}
& \textbf{Category}
& \multicolumn{1}{c}{\textbf{AUC}$\uparrow$}
& \multicolumn{1}{c}{\textbf{ACC}$\uparrow$}
& \multicolumn{1}{c}{\textbf{BACC}$\uparrow$}
& \multicolumn{1}{c}{\textbf{SEN}$\uparrow$}
& \multicolumn{1}{c}{\textbf{SPE}$\uparrow$}
& \multicolumn{1}{c}{\textbf{Macro-F1}$\uparrow$} \\
\midrule
\multicolumn{8}{@{}l}{\textit{Brain-only baselines}} \\
\addlinespace[0.08em]
ConvNeXt~\cite{woo2023convnextv2}
& B/B
& 0.740$\pm$0.039$^*$
& 0.661$\pm$0.051$^*$
& 0.660$\pm$0.049$^*$
& 0.631$\pm$0.102$^*$
& 0.690$\pm$0.165$^*$
& 0.656$\pm$0.050$^*$ \\

3D ViT~\cite{dosovitskiy2021image}
& B/B
& 0.691$\pm$0.063$^*$
& 0.599$\pm$0.061$^*$
& 0.596$\pm$0.064$^*$
& 0.443$\pm$0.267$^*$
& 0.750$\pm$0.155$^*$
& 0.561$\pm$0.130$^*$ \\

Swin-T~\cite{liu2021swin}
& B/B
& 0.693$\pm$0.072$^*$
& 0.619$\pm$0.077$^*$
& 0.618$\pm$0.080$^*$
& 0.551$\pm$0.251$^*$
& 0.684$\pm$0.168$^*$
& 0.601$\pm$0.102$^*$ \\

MiSePyNet~\cite{pan2021misepynet}
& B/B
& 0.935$\pm$0.011
& \underline{0.869$\pm$0.021}
& \underline{0.869$\pm$0.020}
& \underline{0.883$\pm$0.069}
& 0.855$\pm$0.069
& \underline{0.869$\pm$0.021} \\

\addlinespace[0.16em]
\specialrule{\lightrulewidth}{0pt}{0pt}
\addlinespace[0.16em]
\multicolumn{8}{@{}l}{\textit{Privileged-training comparators}} \\
\addlinespace[0.08em]

MOGAD~\cite{zhao2026mogad}
& BG/B
& \underline{0.936$\pm$0.020}
& 0.835$\pm$0.051$^*$
& 0.836$\pm$0.049$^*$
& 0.863$\pm$0.129
& 0.810$\pm$0.174$^*$
& 0.832$\pm$0.055$^*$ \\

BHG/Positional~\cite{li2026tmi_positional}
& BG/B
& 0.883$\pm$0.193
& 0.828$\pm$0.156$^*$
& 0.825$\pm$0.151$^*$
& 0.878$\pm$0.141
& 0.771$\pm$0.387$^*$
& 0.795$\pm$0.224$^*$ \\

\addlinespace[0.18em]
\specialrule{\lightrulewidth}{0pt}{0pt}
\addlinespace[0.18em]
\multicolumn{8}{@{}l}{\textit{Proposed method}} \\
\addlinespace[0.08em]

\rlap{\textcolor{proposedrowgray}{\rule[-\dimexpr\dp\strutbox+2pt\relax]{\textwidth}{\dimexpr\ht\strutbox+\dp\strutbox+4pt\relax}}}Brain-only
& B/B
& 0.909$\pm$0.021
& 0.863$\pm$0.043
& 0.867$\pm$0.043
& 0.853$\pm$0.112$^*$
& \underline{0.881$\pm$0.114}
& 0.861$\pm$0.044 \\

\rlap{\textcolor{proposedrowgray}{\rule[-\dimexpr\dp\strutbox+2pt\relax]{\textwidth}{\dimexpr\ht\strutbox+\dp\strutbox+4pt\relax}}}\textbf{\bvg{}}
& BG/B
& \textbf{0.956$\pm$0.022}
& \textbf{0.908$\pm$0.039}
& \textbf{0.907$\pm$0.039}
& \textbf{0.903$\pm$0.105}
& \textbf{0.911$\pm$0.097}
& \textbf{0.903$\pm$0.041} \\

\bottomrule
\end{tabular*}
\end{table*}

\subsubsection{Evaluation Metrics}
The primary evaluation task was CN/MCI classification on the private cohort, with MCI treated as the positive class. Following a five-fold diagnosis-stratified protocol, approximately 70\%, 10\%, and 20\% of the data were assigned to training, validation, and testing, respectively. All methods used identical disjoint partitions, with model selection and classification thresholds determined on the validation set. For all methods, inference was performed using brain PET alone.
	We report the area under the receiver operating characteristic curve (AUC), accuracy (ACC), balanced accuracy (BACC), sensitivity (SEN), specificity (SPE), and macro-F1, expressed as mean$\pm$SD across the five test folds. For exploratory ablation analyses, statistical uncertainty was assessed using 5,000 fold- and diagnosis-stratified bootstrap resamples, with Holm correction for multiple comparisons.

\subsection{Comparison With State-of-the-Arts}
\subsubsection{Primary Private-Cohort Evaluation}
To determine whether transferred cross-organ information improves brain-only diagnosis, we first evaluate \bvg{} on the primary private-cohort CN/MCI task. As shown in Table~\ref{tab:private_comparison}, \bvg{} reaches an AUC of $0.881$ and BACC of $0.819$, representing relative gains of $6.0\%$ and $6.8\%$ over its protocol-matched brain-only counterpart. Statistical significance is assessed using two-sided paired student $t$-tests \cite{efron2021computer} across the five matched outer folds with Holm correction for multiple comparisons. A superscript $*$ denotes adjusted $p<0.05$. Among competing methods, ConvNeXt and Swin-T share the highest AUC at $0.837$. Importantly, \bvg{} obtains these gains while using brain PET alone during private-cohort adaptation and testing, whereas the privileged-training comparators additionally require paired private brain--gut PET during their teacher stages.

\begin{figure}[!t]
    \centering
    \includegraphics[width=0.95\columnwidth]{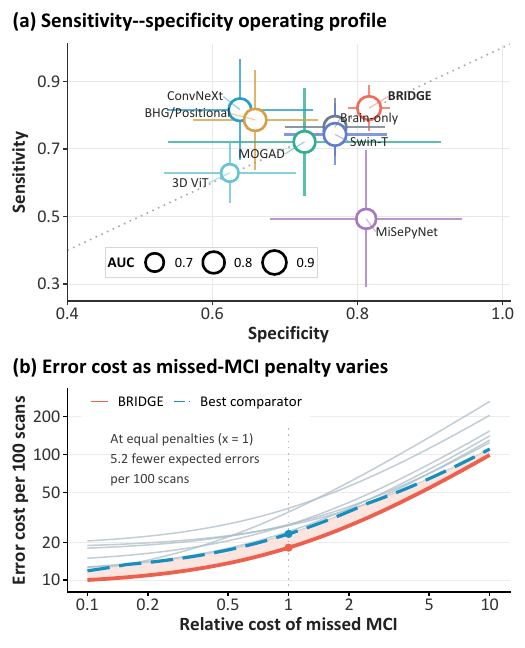}
    \vspace{-10pt}
    \caption{\textbf{Private-cohort operating-point analysis.} (a) Sensitivity and specificity across methods. Markers show five-fold means, error bars show SD, and marker size denotes AUC. The diagonal indicates equal sensitivity and specificity. (b) Expected error cost per 100 scans from five-fold mean sensitivity and specificity and cohort proportions, plotted against the missed-MCI penalty relative to false alarms. Red denotes \bvg{}, gray the comparators, and the navy dashed curve their lowest cost. Both axes are logarithmic, with lower error costs indicating better performance and the dotted line at 1 marking equal penalties.}
    \label{fig:private_heldout_evaluation}
    \vspace{-13pt}
\end{figure}

\subsubsection{Source-Domain Evaluation}
As a complementary source-domain evaluation, we compare the same method families on ADNI CN/AD diagnosis. As shown in Table~\ref{tab:adni_cnad_folds}, \bvg{} achieves an AUC of $0.956$, corresponding to a $5.2\%$ relative gain over its matched brain-only model. Its BACC reaches $0.907$, compared with $0.869$ for MiSePyNet, while the leading privileged-training comparator attains an AUC of $0.936$. Statistical significance is assessed using the same paired testing protocol and significance criterion described above. Together with the private-cohort results, these findings support the applicability of the proposed framework to both CN/AD and CN/MCI diagnosis across public and private cohorts.

\begin{table*}[t]
	\centering
	\caption{Ablation study on private-cohort CN/MCI classification. Results are mean$\pm$SD over five held-out folds. Component-level variants modify individual modules, whereas objective-level variants modify individual loss terms.}
	\label{tab:private_ablation}
	\setlength{\tabcolsep}{4.2pt}
	\renewcommand{\arraystretch}{1.3}
	\small
	\begin{tabular*}{\textwidth}{@{\extracolsep{\fill}}lcccccc@{}}
		\toprule
		\textbf{Variant}
		& \textbf{AUC}$\uparrow$
		& \textbf{ACC}$\uparrow$
		& \textbf{BACC}$\uparrow$
		& \textbf{SEN}$\uparrow$
		& \textbf{SPE}$\uparrow$
		& \textbf{Macro-F1}$\uparrow$ \\
		\midrule
		
		\textbf{Full \bvg{}}
		& \textbf{0.881$\pm$0.050}
		& \textbf{0.818$\pm$0.028}
		& \textbf{0.819$\pm$0.027}
		& \textbf{0.821$\pm$0.067}
		& \textbf{0.816$\pm$0.029}
		& \textbf{0.815$\pm$0.027} \\
		
		\addlinespace[0.16em]
		\specialrule{\lightrulewidth}{0pt}{0pt}
		\addlinespace[0.16em]
		\multicolumn{7}{@{}l}{\textit{Component-level ablations}} \\
		\addlinespace[0.10em]
		
		\quad Brain-only
		& 0.831$\pm$0.030
		& 0.766$\pm$0.032
		& 0.767$\pm$0.032
		& 0.764$\pm$0.086
		& 0.769$\pm$0.069
		& 0.766$\pm$0.031 \\
		
		\quad Without evidence anchors
		& 0.875$\pm$0.043
		& 0.802$\pm$0.028
		& 0.802$\pm$0.028
		& 0.814$\pm$0.073
		& 0.790$\pm$0.057
		& 0.802$\pm$0.027 \\
		
		\quad Without latent transport
		& 0.845$\pm$0.054
		& 0.774$\pm$0.031
		& 0.774$\pm$0.031
		& 0.757$\pm$0.030
		& 0.790$\pm$0.044
		& 0.774$\pm$0.031 \\
		
		\quad Fixed residual admission ($\alpha=1$)
		& 0.876$\pm$0.027
		& 0.791$\pm$0.008
		& 0.791$\pm$0.008
		& 0.800$\pm$0.041
		& 0.783$\pm$0.037
		& 0.791$\pm$0.008 \\
		
		\quad Without counterfactual-confidence supervision
		& 0.865$\pm$0.033
		& 0.781$\pm$0.031
		& 0.781$\pm$0.031
		& 0.771$\pm$0.082
		& 0.790$\pm$0.047
		& 0.780$\pm$0.031 \\
		
		\addlinespace[0.16em]
		\specialrule{\lightrulewidth}{0pt}{0pt}
		\addlinespace[0.16em]
		\multicolumn{7}{@{}l}{\textit{Objective-level ablations}} \\
		\addlinespace[0.10em]
		
		\quad Without matched-gain penalty
		& 0.861$\pm$0.016
		& 0.788$\pm$0.022
		& 0.783$\pm$0.022
		& 0.811$\pm$0.056
		& 0.754$\pm$0.037
		& 0.787$\pm$0.022 \\
		
		\quad Without gate-supervision loss
		& 0.872$\pm$0.031
		& 0.784$\pm$0.024
		& 0.784$\pm$0.024
		& 0.793$\pm$0.069
		& 0.775$\pm$0.047
		& 0.784$\pm$0.024 \\
		
		\quad Without residual regularization
		& 0.854$\pm$0.024
		& 0.773$\pm$0.026
		& 0.774$\pm$0.027
		& 0.793$\pm$0.047
		& 0.754$\pm$0.050
		& 0.773$\pm$0.027 \\
		
		\quad Focal classification loss ($\gamma=2$)
		& 0.871$\pm$0.056
		& 0.781$\pm$0.038
		& 0.781$\pm$0.038
		& 0.764$\pm$0.093
		& 0.798$\pm$0.085
		& 0.780$\pm$0.038 \\		
		\bottomrule
	\end{tabular*}
\end{table*}

\subsubsection{Operating-Point Analysis}
To further examine operating behavior, Fig.~\ref{fig:private_heldout_evaluation}(a) shows the sensitivity--specificity profile of each method. Consistent with Table~\ref{tab:private_comparison}, \bvg{} is the only method that improves all six metrics over its protocol-matched brain-only model, reaching sensitivity of $0.821$ and specificity of $0.816$. ConvNeXt approaches the same sensitivity but has substantially lower specificity, whereas MiSePyNet shows the opposite imbalance. Due to the potentially unequal costs of missed MCI and false alarms, Fig.~\ref{fig:private_heldout_evaluation}(b) tests this advantage across relative penalties. \bvg{} retains the lowest expected cost, yielding $5.2$ fewer expected errors per 100 scans than the best comparator at equal penalties. At unchanged validation-selected thresholds, this pattern suggests that brain--gut transfer improves the balance between MCI detection and false alarms.

\subsection{Ablation and Transport Analysis}
\begin{figure}[!t]
	\centering
	\includegraphics[width=\columnwidth]{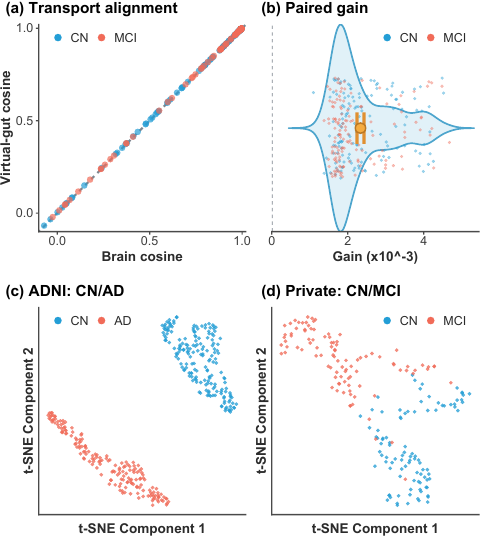}
	\vspace{-15pt}
	\caption{\textbf{Cross-organ transport alignment and task-specific feature geometry.} (a) Cosine similarities of cerebral and transported virtual-gut latents to matched measured-gut latents ($n=278$). The dashed line marks equality. (b) Paired gains after transport, with the point and bar indicating the mean and stratified-bootstrap 95\% CI, respectively. (c) ADNI CN/AD fold-1 training features (182 CN, 174 AD). (d) Private-cohort CN/MCI fold-1 training features (96 CN, 98 MCI).}
	\label{fig:task_specific_tsne}
	\vspace{-10pt}
\end{figure}

\subsubsection{Component-Level Ablation}
To assess the roles of the major components, we compare removal and replacement variants under the same private-cohort protocol, with the brain-only variant providing the matched reference for assessing the added virtual-gut pathway. As shown in Table~\ref{tab:private_ablation}, removing latent transport produces the largest AUC decline among single-component variants, from $0.881$ to $0.845$, and lowers sensitivity from $0.821$ to $0.757$. Removing counterfactual-confidence supervision lowers BACC from $0.819$ to $0.781$, whereas fixed residual admission reduces specificity from $0.816$ to $0.783$. Evidence-anchor removal has a more modest effect, lowering BACC from $0.819$ to $0.802$. Overall, the results suggest complementary contributions from transport, semantic guidance, and adaptive residual fusion. 

\subsubsection{Objective-Level Ablation}
To assess the contribution of the optimization objective, we examine objective-level variants while keeping the full architecture unchanged. Table~\ref{tab:private_ablation} shows that removing residual regularization lowers BACC from $0.819$ to $0.774$ and specificity from $0.816$ to $0.754$. Removing gate-supervision loss reduces BACC by $4.3\%$ relative to the full objective, while focal loss yields an AUC of $0.871$ but sensitivity of only $0.764$. The complete objective therefore leads across all six mean metrics under the current protocol. Together, these ablations suggest that matched-gain penalization, gate supervision, and residual regularization contribute in complementary ways to a balanced operating profile.

\subsubsection{Cross-Organ Transport Analysis}
Beyond classification performance, we assess whether the frozen transport moves cerebral representations toward subject-matched measured-gut counterparts. As shown in Fig.~\ref{fig:task_specific_tsne}(a) and (b), mean cosine similarity to the measured-gut latent increases from 0.8467 to 0.8491 after transport. All 278 paired examinations show positive gains (mean $\Delta=0.00233$; stratified-bootstrap 95\% CI, 0.00224--0.00243), corresponding to a 0.276\% relative increase in mean cosine similarity. The modest magnitude is consistent with a subtle, subject-specific representation-level adjustment rather than a large transformation or explicit anatomical reconstruction. Measured gut PET provides a subject-matched reference for transport alignment, while adaptation and inference rely on brain PET alone.

\subsection{Representation and Interpretability Analysis}
\subsubsection{Task-Specific Feature Geometry}
To characterize the organization of the learned diagnostic features, we visualize task-specific t-SNE projections of gated \bvg{} representations from the ADNI CN/AD and private CN/MCI training subsets in Fig.~\ref{fig:task_specific_tsne}(c) and (d). The embeddings are generated independently for the two diagnostic tasks and are used only to examine within-task representation geometry and class organization. The ADNI CN/AD representation shows comparatively clear group separation, whereas the private CN/MCI representation exhibits a broader transition between diagnostic categories. Neighborhood label agreement remains high ($>0.90$) in both tasks, indicating diagnosis-related local organization despite the subtler CN/MCI distinction. Together, these visualizations illustrate diagnostic group structure in both training cohorts, complementing the quantitative classification results.

\subsubsection{Regional Attribution Analysis}
\begin{figure}[!t]
	\centering
	\includegraphics[width=\columnwidth]{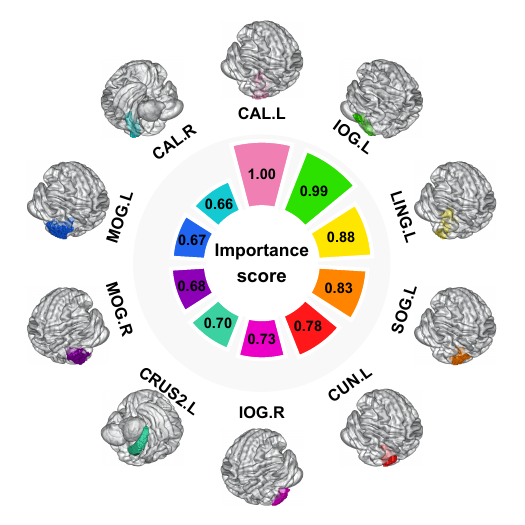}
	\vspace{-15pt}
	\caption{\textbf{Out-of-fold regional attribution on the private cohort.} Circular bars show the ten AAL3 regions with the highest mean normalized positive integrated-gradient attribution to the final MCI logit, with scores rescaled to the top-ranked region. Surrounding surfaces localize the corresponding atlas parcels. Results aggregate all 278 examinations and reflect model-attributed diagnostic evidence rather than causal anatomy or imaging biomarkers.}
	\label{fig:roi_attribution}
	\vspace{-10pt}
\end{figure}

To localize the cerebral evidence underlying private-cohort predictions, we aggregate out-of-fold integrated-gradient attribution for the final MCI logit within AAL3 regions and summarize the most influential regions in Fig.~\ref{fig:roi_attribution}. The attribution is concentrated in posterior cerebral regions, with the left calcarine and inferior occipital cortices showing the strongest contributions and additional involvement of lingual, cuneal, and bilateral occipital regions. Bilateral calcarine, inferior occipital, and middle occipital parcels, together with lingual and cuneal regions, form a coherent posterior visual-cortical grouping, while left cerebellar Crus II contributes outside this cortical cluster. This spatial organization links model attribution to identifiable anatomical regions, clarifying how regional cerebral responses contribute to the model's MCI classification decisions.

\subsubsection{Brain--Gut Grad-CAM Visualization}
\begin{figure}[!t]
	\centering
	\includegraphics[width=\columnwidth]{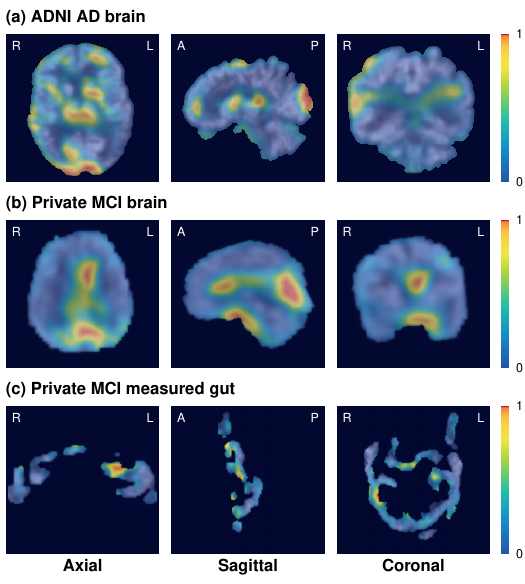}
	\vspace{-15pt}
	\caption{\textbf{Illustrative brain--gut Grad-CAM maps.} (a) ADNI AD brain PET, with attribution to the AD-class diagnostic logit. (b) Private-cohort MCI brain PET, with attribution to the MCI-class diagnostic logit. (c) Matched measured-gut PET from the same private examination, probed post hoc through the trained virtual-gut residual head. Columns show axial, sagittal, and coronal views.}
	\label{fig:brain_gut_gradcam}
\end{figure}

To examine the spatial responses underlying individual predictions, Fig.~\ref{fig:brain_gut_gradcam} presents an ADNI AD brain examination and a paired private MCI brain--gut examination in three orthogonal views. Both brain maps highlight posterior responses, while the MCI example also shows an elongated central pattern in the axial view. These distributions illustrate the cerebral regions contributing to the model's predictions. Complementing the brain maps, the measured-gut map from the same private case highlights localized responses of the trained residual pathway, providing a post-hoc anatomical reference for interpreting gut-related evidence. Together, these maps complement the quantitative results with localized cerebral and gut-residual responses, clarifying \bvg{}'s brain-only diagnostic framework.

\section{Conclusion}\label{CD}
This study presents BRIDGE, a framework that transfers metabolic knowledge from paired whole-body PET to cognitive impairment diagnosis using brain PET alone. BRIDGE integrates conditional brain-to-gut latent transport, cerebral representation learning guided by disease evidence, and counterfactual-guided residual fusion to model subject-specific correspondence between brain and gut representations and adaptively regulate the diagnostic contribution of virtual gut evidence without requiring measured gut PET at inference. By separating cross-organ representation learning from downstream diagnostic supervision, the framework enables paired whole-body PET and labeled brain PET from distinct cohorts to contribute to a unified diagnostic model. Experiments across public and private cohorts demonstrate that BRIDGE achieves AUCs of 0.881 and 0.956 for CN/MCI and CN/AD diagnosis, representing relative improvements of 6.0\% and 5.2\%, respectively, over its corresponding brain PET models evaluated under the same protocol, while outperforming existing cross-organ approaches. Ablation and representation analyses support the complementary roles of latent transport, semantic guidance, and adaptive residual fusion. These findings demonstrate the feasibility of learning transferable brain–gut metabolic correspondence from heterogeneous data resources and incorporating it into brain-only diagnosis. Future work will assess the robustness and generalizability of BRIDGE across independent cohorts and imaging protocols.

\bibliographystyle{IEEEtran}
\bibliography{references}

\end{document}